\documentclass[11pt,a4paper]{article}
\pdfoutput=1
\usepackage[hyperref]{emnlp-ijcnlp-2019}
\usepackage{times}
\usepackage{latexsym}
\usepackage{amsfonts}
\usepackage{amsmath}
\usepackage{paralist}
\usepackage{multirow}
\usepackage{hhline}

\usepackage{url}

\aclfinalcopy % Uncomment this line for the final submission

\title{WikiCREM: A Large Unsupervised Corpus for Coreference Resolution}

\author{Vid Kocijan\textsuperscript{1} \ \  Oana-Maria Camburu\textsuperscript{1,2} \ \  Ana-Maria Cre\c{t}u\textsuperscript{3} \ \  Yordan Yordanov\textsuperscript{1}\\ \textbf{Phil Blunsom\textsuperscript{1,4} \ \  Thomas Lukasiewicz\textsuperscript{1,2}} \\
  \textsuperscript{1}University of Oxford, UK \ \ 
  \textsuperscript{2}Alan Turing Institute, London, UK \\
  \textsuperscript{3}Imperial College London, UK  \ \
  \textsuperscript{4}DeepMind, London, UK \\
  \texttt{firstname.lastname@cs.ox.ac.uk, a.cretu@imperial.ac.uk}}%

\date{}
\begin{document}
\maketitle

\begin{abstract}
    Pronoun resolution is a major area of natural language understanding. However, large-scale training sets are still scarce, since manually labelling data is costly.
    In this work, we introduce \textsc{WikiCREM} (Wikipedia CoREferences Masked) a large-scale, yet accurate dataset of pronoun disambiguation instances. 
    We use a language-model-based approach for pronoun resolution in combination with our \textsc{WikiCREM} dataset.
    We compare a series of models on a collection of diverse and challenging coreference resolution problems, where we match or outperform previous state-of-the-art approaches on $6$ out of $7$ datasets, such as \textsc{Gap}, \textsc{Dpr}, \textsc{Wnli}, \textsc{Pdp}, \textsc{WinoBias}, and \textsc{WinoGender}. We release our model to be used off-the-shelf for solving pronoun disambiguation.

\end{abstract}
\section{Introduction}
Pronoun resolution, also called coreference or ana\-phora resolution, is a natural language processing (NLP) task, which aims to link the pronouns with their referents.
This task is of crucial importance in various other NLP tasks, such as information extraction \cite{AnaphoraforIE} and machine translation \cite{AnaphoraforMT}.
Due to its importance, pronoun resolution has seen a series of different approaches, such as rule-based systems \cite{CoreferenceRules} and end-to-end-trained neural models \cite{CoreferenceNeural,ReferentialReader}.
However, the recently released dataset \textsc{Gap} \cite{Gap} shows that most of these solutions perform worse than na\"ive baselines when the answer cannot be deduced from the syntax.
Addressing this drawback is difficult, partially due to the lack of large-scale challenging datasets needed to train the data-hungry neural models.

As observed by \citeauthor{WinogradGoogle}~(\citeyear{WinogradGoogle}), language models are a natural approach to pronoun resolution, by selecting the replacement for a pronoun that forms the sentence with highest probability.
Additionally, language models have the advantage of being pre-trained on a large collection of unstructured text and then fine-tuned on a specific task using much less training data. This procedure has obtained state-of-the-art results on a series of natural language understanding tasks \cite{Bert}.

In this work, we address the lack of large training sets for pronoun disambiguation by introducing a large dataset that can be easily extended.
To generate this dataset, we find passages of text where a personal name appears at least twice and mask one of its non-first occurrences. To make the disambiguation task more challenging, we also ensure that at least one other distinct personal name is present in the text in a position before the masked occurrence. We instantiate our method on English Wikipedia and generate the Wikipedia Co-REferences Masked (\textsc{WikiCREM}) dataset with $2.4$M examples, which we make publicly available for further usage\footnote{The code can be found at \url{https://github.com/vid-koci/bert-commonsense}.\\ The dataset and the models can be obtained from \url{https://ora.ox.ac.uk/objects/uuid:c83e94bb-7584-41a1-aef9-85b0e764d9e3}}. We show its value by using it to fine-tune the \textsc{Bert} language model \cite{Bert} for pronoun resolution.

To show the usefulness of our dataset, we train several models that cover three real-world scenarios: (1) when the target data distribution is completely unknown, (2) when training data from the target distribution is available, and (3) the transductive scenario, where the unlabeled test data is available at the training time.
We show that fine-tuning BERT with \textsc{WikiCREM} consistently improves the model in each of the three scenarios, when evaluated on a collection of $7$ datasets.
For example, we outperform the state-of-the-art approaches on \textsc{Gap} \cite{Gap}, \textsc{Dpr} \cite{DPR}, and \textsc{Pdp} \cite{WSC2016} by $5.9\%$, $8.4\%$, and $12.7\%$, respectively.
%For example, we achieve an $F_1$-score of $76.4$ on the \textsc{Gap} \cite{Gap} test set ($4.3\%$ absolute improvement over the state-of-the-art), an accuracy $83.5\%$ on the \textsc{Dpr} dataset \cite{DPR} ($7.1\%$ absolute improvement), $73.3\%$ on \textsc{Wnli} \cite{Glue} ($0.7\%$ absolute improvement), $85\%$ on \textsc{Pdp} \cite{WSC2016} ($11\%$ absolute improvement).
Additionally, models trained with \textsc{WikiCREM} show increased performance and reduced bias on the gender diagnostic datasets \textsc{WinoGender} \cite{WinoGender} and \textsc{WinoBias} \cite{WinoBias}.
\section{Related Work}

There are several large and commonly used benchmarks for coreference resolution, such as \cite{Conll12,ACLCoref,WikiCoref}. However, \citeauthor{Gap}~(\citeyear{Gap}) argue that a high performance on these datasets does not correlate with a high accuracy in practice, because examples where the answer cannot be deduced from the syntax (we refer to them as \textit{hard pronoun resolution}) are underrepresented.
Therefore, several hard pronoun resolution datasets have been introduced \cite{Gap,DPR,WinoGender,WSC2016,WinoBias,knowref}. However, they are all relatively small, often created only as a test set.

Therefore, most of the pronoun resolution models that address hard pronoun resolution rely on little \cite{ReferentialReader} or no training data, via unsupervised pre-training \cite{WinogradGoogle, GPT2}.
Another approach involves using external knowledge bases \cite{KnowledgeHunter, MarkerPassing}, however, the accuracy of these models still lags behind that of the aforementioned pre-trained models.

A similar approach to ours for unsupervised data generation and language-model-based evaluation has been recently presented in our previous work \cite{MaskedWiki}.
We generated \textsc{MaskedWiki}, a large unsupervised dataset created by searching for repeated occurrences of nouns. % and then additionally filter it using a language model to reduce the number of easy examples.
However, training on \textsc{MaskedWiki} on its own is not always enough and sometimes makes a difference only in combination with additional training on the \textsc{Dpr} dataset (called \textsc{Wscr}) \cite{DPR}.
In contrast, \textsc{WikiCREM} brings a much more consistent improvement over a wider range of datasets, strongly improving models' performance even when they are not fine-tuned on additional data. As opposed to our previous work \cite{MaskedWiki}, we evaluate models on a larger collection of test sets, showing the usefulness of \textsc{WikiCREM} beyond the Winograd Schema Challenge.

Moreover, a manual comparison of \textsc{WikiCREM} and \textsc{MaskedWiki} \cite{MaskedWiki} shows a significant difference in the quality of the examples.
We annotated $100$ random examples from \textsc{MaskedWiki} and \textsc{WikiCREM}. In \textsc{MaskedWiki}, we looked for examples where masked nouns can be replaced with a pronoun, and only in $7$ examples, we obtained a natural-sounding and grammatically correct sentence. In contrast, we 
estimated that $63\%$ of the annotated examples in \textsc{WikiCREM} form a natural-sounding sentence when the appropriate pronoun is inserted, showing that \textsc{WikiCREM} consists of examples that are much closer to the target data.
We highlight that pronouns are not actually inserted into the sentences and thus none of the examples sound unnatural.
This analysis was performed to show that \textsc{WikiCREM} consists of examples with data distribution closer to the target tasks than \textsc{MaskedWiki}.

%TODO: expand?

%The GAP dataset was recently introduced to mitigate the above problems and measure the gender bias of the off-the-shelf models.
%It is a collection of $4,454$ paragraphs from Wikipedia with ambiguous pronouns.
%The passages were randomly sampled from the Wikipedia biography pages and manually annotated.
%The dataset specifically focuses on the resolution of personal pronouns that refer to human names.
%It contains 1:1 ratio between masculine and feminine pronouns and checks the models for their gender-bias by looking at the ratio between performance on masculine and feminine pronouns. 
%%In addition to its overall $F_1$ score, each model is measured on its performance on masculine ($F_{1\mathbf{M}}$) and feminine pronouns ($F_{1\mathbf{F}}$) and bias, defined as the ratio between them $\frac{F_{1\mathbf{F}}}{F_{1\mathbf{M}}}$. 
%The authors introduce a simple but strong baseline that they call Parallelism.
%If the pronoun is the subject or the direct object, they pick the nearest candidate with the same grammatical argument.
%When equipped with additional context from the web-page that the passage was extracted from, this baseline achieved a higher result than previous models \cite{CoreferenceNeural, CoreferenceRules}.
%This baseline was named Parallelism+URL.
%The only model that beats this baseline is the Referential Reader \cite{ReferentialReader}, a GRU-based model with additional external memory cells.

\section{The \textsc{WikiCREM} Dataset}
\label{section-dataset}

%To address the lack of large-scale pronoun resolution datasets, we introduce a large unsupervised corpus of anaphora resolution examples.
In this section, we describe how we obtained \textsc{WikiCREM}.
Starting from English Wikipedia\footnote{\url{https://dumps.wikimedia.org/enwiki/} dump id: enwiki-20181201}, we search for sentences and pairs of sentences with the following properties: at least two distinct personal names appear in the text, and one of them is repeated.
We do not use pieces of text with more than two sentences to collect concise examples only.
Personal names in the text are called ``candidates''.
One non-first occurrence of the repeated candidate is masked, and the goal is to predict the masked name, given the correct and one incorrect candidate.
In case of more than one incorrect candidate in the sentence, several datapoints are constructed, one for each incorrect candidate.

We ensure that the alternative candidate appears before the masked-out name in the text, in order to avoid trivial examples.
%More specifically, we follow the pattern named \textbf{\textsc{FinalPro}} in the work of \citeauthor{Gap}~(\citeyear{Gap}).
Thus, the example is retained in the dataset~if:
\begin{compactenum}
    \item[(a)] the repeated name appears after both candidates, all in a single sentence; or
    \item[(b)] both candidates appear in a single sentence, and the repeated name appears in a sentence directly following.
\end{compactenum}
Examples where one of the candidates appears in the same sentence as the repeated name, while the other candidate does not, are discarded, as they are often too trivial.
%\citeauthor{Gap}~(\citeyear{Gap}) also introduce patterns named \textbf{\textsc{InitialPro}} and \textbf{\textsc{MedialPro}}, where the pronoun (repeated word in our case) appears before or between the candidates, respectively.
%We do not introduce examples with these patterns to the \textsc{WikiCREM} dataset as they do not seem to positively impact the model.

We illustrate the procedure with the following example:

\smallskip
\textit{When asked about \textbf{Adams}' report, \textbf{Powell} found many of the statements to be inaccurate, including a claim that \textbf{Adams} first surveyed an area that was surveyed in 1857 by Joseph C.}

The second occurrence of ``Adams'' is masked.
The goal is to determine which of the two candidates (``Adams'',``Powell'') has been masked out.
The masking process resembles replacing a name with a pronoun, but the pronoun is not inserted to keep the process fully unsupervised and error-free.

%Additionally, it reduces the number of ambiguous examples where nouns are synonyms or refer to the same entity. [In the end I believe this wouldn't actually be an advantage because the "semantic" training wouldn't be hurt since a correct reference is given.]

%We use English Wikipedia as our starting point for generating our \textsc{WikiCREM} dataset while noting that other large scale sources of data could be used to further extend it. %the source of English text because it is a large collection of diverse and grammatically correct text.
%Only passages with at most two sentences were considered to retain concise examples solely.
We used the Spacy Named Entity Recognition library\footnote{\url{https://spacy.io/usage/linguistic-features\#named-entities}} to find the occurrences of names in the text.
The resulting dataset consists of $2,438,897$ samples.
$10,000$ examples are held out to serve as the validation set.
%This dataset is further split into the training set with $2,428,897$ samples and the validation set with $10,000$ samples. 
Two examples from our dataset can be found on Figure~\ref{dataset-example}.

\begin{figure}[h]

\smallskip
\textit{Gina arrives and she is furious with Denise for not protecting Jody from
Kingsley, as \texttt{[MASK]} was meant to be the parent.}

\textbf{Candidates:} Gina, \textbf{Denise}
\smallskip

\textit{When Ashley falls pregnant with Victor's child, Nikki is diagnosed with cancer, causing Victor to leave \texttt{[MASK]}, who secretly has an abortion.}

\textbf{Candidates:} \textbf{Ashley}, Nikki
\smallskip

\caption{\textsc{WikiCREM} examples. Correct answers are given in bold.}
\label{dataset-example}
\end{figure}
We note that our dataset contains hard examples.
To resolve the first example, one needs to understand that Denise was assigned a task and ``meant to be the parent'' thus refers to her.
To resolve the second example, one needs to understand that having an abortion can only happen if one falls pregnant first.
Since both candidates have feminine names, the answer cannot be deduced just on the common co-occurrence of female names and the word ``abortion''.

We highlight that our example generation method, while having the advantage of being unsupervised, also does not give incorrect signals, since we know the ground truth reference. %for the ``pronoun''. 

Even though \textsc{WikiCREM} and \textsc{Gap} both use text from English Wikipedia, they produce differing examples, because their generating processes differ.
In \textsc{Gap}, passages with pronouns are collected and the pronouns are manually annotated, while \textsc{WikiCREM} is generated by masking names that appear in the text.
Even if the same text is used, different masking process will result in different inputs and outputs, making the examples different under the transductive hypothesis.

\paragraph{\textsc{WikiCREM} statistics.}
We analyze our dataset for gender bias.
We use the Gender guesser library\footnote{https://pypi.org/project/gender-guesser/} to determine the gender of the candidates.
To mimic the analysis of pronoun genders performed in the related works \cite{Gap,WinoGender,WinoBias}, we observe the gender of the correct candidates only. 
There were $0.8$M ``male'' or ``mostly\_male'' names and $0.42$M ``female'' or ``mostly\_female'' names, the rest were classified as ``unknown''.
The ratio between female and male candidates is thus estimated around $0.53$ in favour of male candidates.
We will see that this gender imbalance does not have any negative impact on bias,
%and that fine-tuning on \textsc{WikiCREM} often actually reduces the bias, 
as shown in Section~\ref{section-results}.
%Even though the dataset contains many more male than female candidates it seemed to balance the performance on male and female subset of GAP, as shown in Section \ref{section-eval}.

However, our unsupervised generating procedure sometimes yields examples where the correct answer cannot be deduced given the available information, we refer to these as \textit{unsolvable examples}.
To estimate the percentage of unsolvable examples, we manually annotated $100$ randomly selected examples from the \textsc{WikiCREM} dataset.
In order to prevent guessing, the candidates were not visible to the annotators.
For each example, we asked them to state whether it was solvable or not, and to answer the solvable examples.
In $100$ examples, we found $18$ unsolvable examples and achieved $95.1\%$ accuracy on the rest, showing that the annotation error rate is tolerable.
These annotations can be found in  Appendix~\ref{appendix}.

However, as shown in Section~\ref{section-results}, training on \textsc{WikiCREM} alone does not match the performance of training on the data from the target distribution.
The data distribution of \textsc{WikiCREM} differs from the data distribution of the datasets for evaluation.
If we replace the \texttt{[MASK]} token with a pronoun instead of the correct candidate, the resulting sentence sometimes sounds unnatural and would not occur in a human-written text.
On the annotated $100$ examples, we estimated the percentage of natural-sounding sentences to be $63\%$.
While the these sentences are not incorrect, the distribution of the training data differ from the target data.
\section{Model}

We use a simple language-model-based approach to anaphora resolution to show the value of the introduced dataset.
In this section, we first introduce \textsc{Bert} \cite{Bert}, a language model that we use throughout this work.
In the second part, we describe the utilization of \textsc{Bert} and the fine-tuning procedures employed.

\subsection{\textsc{Bert}}

The Bidirectional Encoder Representations from Transformers (\textsc{Bert}) language model is based on the transformer architecture \cite{AttentionAllYouNeed}.
We choose this model due to its strong language-modeling abilities and high performance on several NLU tasks~\cite{Bert}.

\textsc{Bert} is initially trained on two tasks: next sentence prediction and masked token prediction.
In the next sentence prediction task, the model is given two sentences and is asked to predict whether the second sentence follows the first one.
In the masked token prediction task, the model is given text with approximately $15\%$ of the input tokens masked, and it is asked to predict these tokens.
The details of the pre-training procedure can be found in \citeauthor{Bert}~(\citeyear{Bert}).

In this work, we only focus on the masked token prediction. % for fine-tuning the pre-trained model made available by \citeauthor{Bert}~(\citeyear{Bert}).
We use the PyTorch implementation of \textsc{Bert}\footnote{\url{https://github.com/huggingface/pytorch-pretrained-BERT}} and the pre-trained weights for \textsc{Bert}-large released by \citeauthor{Bert}~(\citeyear{Bert}). 

\subsection{Pronoun Resolution with \textsc{Bert}}

This section introduces the procedure for pronoun resolution used throughout this work.
Let $S$ be the sentence with a pronoun that has to be resolved.
Let $\mathbf{a}$ be a candidate for pronoun resolution.
The pronoun in $S$ is replaced with a \texttt{[MASK]} token and used as the input to the model to compute the log-probability $\log\mathbb{P}(\mathbf{a}|S)$.
If $\mathbf{a}$ consists of more than one token, the same number of \texttt{[MASK]} tokens is inserted into $S$, and the log-probability $\log\mathbb{P}(\mathbf{a}|S)$ is computed as the average of log-probabilities of all tokens in $\mathbf{a}$.

The candidate-finding procedures are dataset-specific and are described in Section~\ref{section-eval}.
Given a sentence $S$ and several candidates $\mathbf{a}_1, \ldots, \mathbf{a}_n$, we select the candidate $\mathbf{a}_i$ with the largest $\log\mathbb{P}(\mathbf{a}_i|S)$.

\subsection{Training}
\label{section-train}

When training the model, the setup is similar to testing.
We are given a sentence with a name or a pronoun masked out, together with two candidates.
The goal is to determine which of the candidates is a better fit.
%In some of the examples in the GAP training set, both candidates for pronoun resolution are annotated as incorrect.
%We discard these examples and do not use them during training.
%Hence, we always have a correct and an incorrect candidate.
Let $\mathbf{a}$ be the correct candidate, and $\mathbf{b}$ be an incorrect candidate.
Following our previous work \cite{MaskedWiki} we minimize the negative log-likelihood of the correct candidate, while additionally imposing a max-margin between the log-likelihood of the correct and incorrect terms.
We observe that this combined loss consistently yields better results on validation sets of all experiments than negative log-likelihood or max-margin loss on their own.
\begin{align}\label{loss}
\mathcal{L}=&-\log\mathbb{P}(\mathbf{a}|S)+&\\
+&\alpha\cdot\text{max}(0,\log\mathbb{P}(\mathbf{b}|S)-\log\mathbb{P}(\mathbf{a}|S)+\beta), \nonumber
\end{align}%
where $\alpha$ and $\beta$ are hyperparameters controlling the influence of the max-margin loss term and the margin between the log-likelihood of the correct and incorrect candidates, respectively.

The hyperparameter settings for fine-tuning \textsc{Bert} on \textsc{WikiCREM} were the same as by \citeauthor{Bert}~(\citeyear{Bert}), except for the learning rate and introduced constants $\alpha$ and $\beta$.
For our hyperparameter search, we used learning rate $lr\in\{3\cdot10^{-5}, 1\cdot10^{-5}, 5\cdot10^{-6}, 3\cdot10^{-6}\}$ and hyperparameters $\alpha\in\{5,10,20\}$, $\beta\in\{0.1,0.2,0.4\}$ with grid search.
The hyperparameter search is performed on a subset of \textsc{WikiCREM} with $10^5$ datapoints to reduce the searching time.
We compare the influence of hyperparameters on the validation set of \textsc{WikiCREM} dataset.
The best validation score was achieved with $lr = 1\cdot10^{-5}$, $\alpha=10$, and $\beta=0.2$. 
We used batches of size $64$.

Since \textsc{WikiCREM} is large and one epoch takes around two days even when parallelized on 8 Tesla P100 GPUs, we only fine-tune \textsc{Bert} on \textsc{WikiCREM} for a single epoch.
We note that better results may be achieved with further fine-tuning and improved hyperparameter search.

Fine-tuning on other datasets is performed in the same way as training except for two differences.
Firstly, in fine-tuning, the model is trained for $30$ epochs due to the smaller size of datasets.
Secondly, we do not sub-sample the training set for hyperparameter search.
We validate the model after every epoch, retaining the model that performs best on the \textsc{WikiCREM} validation set.

%\paragraph{Overlap with GAP}
%%No overlap between \textsc{WikiCREM} and GAP test set was found.
%During the training time, \textsc{Bert} has likely seen all sentences that appear in the GAP test set, however, not in the same form.
%During \textsc{Bert} training time $15\%$ of the input tokens were dropped and BERT was asked to predict them.

\section{Evaluation Datasets}
\label{section-datasets}

We now introduce the $7$ datasets that were used to evaluate the models.
We decide not to use the  \textsc{CoNLL2012} and \textsc{WinoCoref} \cite{Conll12,WinoCoref} datasets, because they contain more general coreference examples than just pronouns.
We did not evaluate on the \textsc{Know\-Ref} dataset \cite{knowref}, since it was not yet publicly available at the time of writing.

\paragraph*{\textsc{Gap}.}
\textsc{Gap} \cite{Gap} is a collection of $4,454$ passages from Wikipedia containing ambiguous pronouns. It focuses on the resolution of personal pronouns referring to human names and has a $1:1$ ratio between masculine and feminine pronouns.
%and checks the models for their gender-bias by looking at the ratio between performance on masculine and feminine pronouns.
In addition to the  overall performance on the dataset, each model is evaluated also on its performance on the masculine subset ($F_1^M$), feminine subset ($F_1^F$), and its gender bias ($\frac{F_1^F}{F_1^M}$).
The best performance was exhibited by the Referential Reader \cite{ReferentialReader}, a GRU-based model with additional external memory cells.

%In addition to its overall $F_1$ score, each model is measured on its performance on masculine ($F_{1\mathbf{M}}$) and feminine pronouns ($F_{1\mathbf{F}}$) and bias, defined as the ratio between them $\frac{F_{1\mathbf{F}}}{F_{1\mathbf{M}}}$. 
%The authors introduce a simple but strong baseline that they call Parallelism.
%If the pronoun is the subject or the direct object, they pick the nearest candidate with the same grammatical argument.
%When equipped with additional context from the web-page that the passage was extracted from, this baseline achieved a higher result than state-of-the-art models for co-reference resolution on the \textsc{CoNLL2012} dataset \cite{CoreferenceNeural, CoreferenceRules}.
%This baseline was named Parallelism+URL.

For each example, two candidates are given with the goal of determining whether they are the referent.
In approximately $10\%$ of the training examples, none of the candidates are correct.
When training on the \textsc{Gap} dataset, we discard such examples from the training set.
We do not discard any examples from the validation or test set.

%When testing the model, the model does not have access to the candidates in questions, but has to be capable of finding them on its own.
When testing the model, we use the Spacy NER library to find all candidates in the sentence. Since the \textsc{Gap} dataset mainly contains examples with human names, we only retain named entities with the tag \textsc{Person}.
We observe that in $18.5\%$ of the test samples, the  Spacy NER library fails to extract the candidate in question, making the answer for that candidate ``FALSE'' by default, putting our models at disadvantage.
Because of this, $7.25\%$ of answers are always false negatives, and $11.25\%$ are always true negatives, regardless of the model.
Taking this into account, we compute that the maximal $F_1$-score achievable by our models is capped at $91.1\%$.

We highlight that, when evaluating our models, we are stricter than previous approaches \cite{ReferentialReader,Gap}.
While they count the answer as ``correct'' if the model returns a substring of the correct answer, we only accept the full answer.
The aforementioned models return the exact location of the correct candidate in the input sentence, while our approach does not.
This strictness is necessary, because a substring of a correct answer could be a substring of several answers at once, making it ambiguous. 

\paragraph*{\textsc{Wsc}.}
The Winograd Schema Challenge \cite{WinogradSchema} is a hard pronoun resolution challenge inspired by the example from \citeauthor{WinogradUnderstandingNL}~(\citeyear{WinogradUnderstandingNL}):

\smallskip 
\noindent\textit{The city councilmen refused the demonstrators a permit because \textbf{they} [feared/advocated] violence.}

\noindent\textbf{Question: } Who [feared/advocated] violence?

\noindent\textbf{Answer: } the city councilmen / the demonstrators

\smallskip 

A change of a single word in the sentence changes the referent of the pronoun, making it very hard to resolve.
An example of a Winograd Schema must meet the following criteria \cite{WinogradSchema}:
\begin{compactenum}
  \item Two entities appear in the text.
  \item A pronoun or a possessive adjective appears in the sentence and refers to one of the entities. It would be grammatically correct if it referred to the other entity.
  \item The goal is to find the referent of the pronoun or possessive adjective.
  \item The text contains a ``special word''. When switched for the ``alternative word'', the sentence remains grammatically correct, but the referent of the pronoun changes.
\end{compactenum}
The Winograd Schema Challenge is specifically made up from challenging examples that require commonsense reasoning for resolution and should not be solvable with statistical analysis of co-occurence and association.

We evaluate the models on the collection of $273$ problems used for the 2016 Winograd Schema Challenge \cite{WSC2016}, also known as \textsc{Wsc273}.
%This collection of problems is often referred to as \textsc{Wsc273}.
The best known approach to this problem uses the \textsc{Bert} language model, fine-tuned on the \textsc{Dpr} dataset \cite{MaskedWiki}.

\paragraph*{\textsc{Dpr}.} 
The Definite Pronoun Resolution (\textsc{Dpr}) corpus \cite{DPR} is a collection of problems that resemble the Winograd Schema Challenge.
The criteria for this dataset have been relaxed, and it contains examples that might not require commonsense reasoning or examples where the ``special word'' is actually a whole phrase.
We remove $6$ examples in the \textsc{Dpr} training set that overlap with the \textsc{Wsc} dataset.
The dataset was constructed manually and consists of $1316$ training and $564$ test samples after we removed the overlapping examples.
The best result on the dataset was reported by \citeauthor{WinoCoref}~(\citeyear{WinoCoref}) using external knowledge sources and integer linear programming.

\paragraph*{\textsc{Pdp}.} 
The Pronoun Disambiguation Problem (\textsc{Pdp}) is a small collection of $60$ problems that was used as the first round of the Winograd Schema Challenge in $2016$ \cite{WSC2016}.
Unlike \textsc{Wsc}, the examples do not contain a ``special word'', however, they do require commonsense reasoning to be answered.
The examples were manually collected from books.
Despite its small size, there have been several attempts at solving this challenge \cite{MarkerPassing,WinogradGoogle}, the best result being held by the Marker Passing algorithm \cite{MarkerPassing}.

\paragraph*{\textsc{Wnli}.}
The Winograd Natural Language Inference (\textsc{Wnli}) is an inference task inspired by the Winograd Schema Challenge and is one of the $9$ tasks on the \textsc{Glue} benchmark \cite{Glue}.
\textsc{Wnli} examples are obtained by rephrasing Winograd Schemas. % into inference problems.
The Winograd Schema is given as the ``premise''. A ``hypothesis'' is constructed by repeating the part of the premise with the pronoun and replacing the pronoun with one of the candidates.
The goal is to classify whether the hypothesis follows from the premise.

A \textsc{Wnli} example obtained by rephrasing one of the \textsc{Wsc} examples looks like this:

\smallskip 
\noindent\textbf{Premise: }\textit{The city councilmen refused the demon\-strators a permit because they feared violence.}

\noindent\textbf{Hypothesis: } \textit{The demonstrators feared violence.}

\noindent\textbf{Answer: } true / \textbf{false}

\smallskip 

The \textsc{Wnli} dataset is constructed manually.
Since the \textsc{Wnli} training and validation sets overlap with \textsc{Wsc}, we use the \textsc{Wnli} test set only.
The test set of \textsc{Wnli} comes from a separate source and does not overlap with any other dataset.

The currently best approach transforms examples back into the Winograd Schemas and solves them as a coreference problem \cite{MaskedWiki}.
Following our previous work \cite{MaskedWiki}, we reverse the process of example generation in the same way.
We automatically detect which part of the premise has been copied to construct the hypothesis.
This locates the pronoun that has to be resolved, and the candidate in question.
All other nouns in the premise are treated as alternative candidates.
We find nouns in the premise with the Stanford POS tagger \cite{StanfordPOS}.

\paragraph*{\textsc{WinoGender}.}
\textsc{WinoGender} \cite{WinoGender} is a dataset that follows the \textsc{Wsc} format and is aimed to measure gender bias.
One of the candidates is always an occupation, while the other is a participant, both selected to be gender neutral.
Examples intentionally contain occupations with strong imbalance in the gender ratio.
%Each occupation is contained in $2$ templates, which are then used to construct several examples.
Participant can be replaced with the neutral ``someone'',  and three different pronouns (he/she/they) can be used.
%These $120$ templates are thus used to construct $120\cdot 2\cdot 3=720$ examples.
The aim of this dataset is to measure how the change of the pronoun gender affects the accuracy of the model.

Our models mask the pronoun and are thus not affected by the pronoun gender.
They exhibit no bias on this dataset by design.
We mainly use this dataset to measure the accuracy of different models on the entire dataset.
According to \citeauthor{WinoGender}~(\citeyear{WinoGender}), the best performance is exhibited by \citeauthor{BaselineWinoGender}~(\citeyear{BaselineWinoGender}) when used on the male subset of the dataset.
We use this result as the baseline.

\paragraph*{\textsc{WinoBias}.}
Similarly to the \textsc{WinoGender} dataset, \textsc{WinoBias} \cite{WinoBias} is a \textsc{Wsc}-inspired dataset that measures gender bias in the coreference resolution algorithms.
Similarly to \textsc{WinoGender}, it contains instances of occupations with high gender imbalance.
It contains $3,160$ examples of Winograd Schemas, equally split into validation and test set. The test set examples are split into $2$ types, where examples of type $1$ are ``harder'' and should not be solvable using the analysis of co-occurrence, and examples of type $2$ are easier.
Additionally, each of these subsets is split into anti-stereotypical and pro-stereotypical subsets, depending on whether the gender of the pronoun matches the most common gender in the occupation.
The difference in performance between pro- and anti-stereotypical examples shows how biased the model is.
The best performance is exhibited by \citeauthor{CoreferenceNeural}~(\citeyear{CoreferenceNeural}) and \citeauthor{BaselineWinoGender}~(\citeyear{BaselineWinoGender}), as reported by \citeauthor{WinoBias}~(\citeyear{WinoBias}).

\section{Evaluation}
\label{section-eval}

We quantify the impact of \textsc{WikiCREM} on the introduced datasets.

\subsection{Experiments}
\label{section-experiments}
We train several different models to evaluate the contribution of the \textsc{WikiCREM} dataset in different real-world scenarios.
In \textbf{Scenario A}, no information of the target distribution is available.
In \textbf{Scenario B}, the distribution of the target data is known and a sample of training data from the target distribution is available. %This sample can be used for additional fine-tuning.
Finally, \textbf{Scenario C} is the transductive scenario where the unlabeled test samples are known in advance.
All evaluations on the GAP test-set are considered to be Scenario C, because \textsc{Bert} has been pre-trained on the English Wikipedia and has thus seen the text in the \textsc{Gap} dataset at the pre-training time.

We describe the evaluated models below.
\paragraph*{\textsc{Bert}.} This model, pretrained by \citeauthor{Bert}~(\citeyear{Bert}), is the starting point for all models and serves as the soft baseline for Scenario A.

\paragraph*{\textsc{Bert\_WikiRand}.} This model serves as an additional baseline for Scenario A and aims to eliminate external factors that might have worked against the performance of \textsc{Bert}. To eliminate the effect of sentence lengths, loss function, and the percentage of masked tokens during the training time, we generate the \textsc{RandomWiki} dataset.
It consists of random passages from Wikipedia and has the same sentence-length distribution and number of datapoints as \textsc{WikiCREM}. However, the masked-out word from the sentence is selected randomly, while the alternative candidate is selected randomly from the vocabulary. \textsc{Bert} is then trained on this dataset in the same way as \textsc{Bert\_WikiCREM}, as described in Section~\ref{section-train}.

\paragraph*{\textsc{Bert\_WikiCREM}.} \textsc{Bert}, additionally train\-ed on \textsc{WikiCREM}. Its evaluation on non-\textsc{Gap} datasets serves as the evaluation of \textsc{WikiCREM} under Scenario A.

\paragraph*{\textsc{Bert\_Dpr}.} \textsc{Bert}, fine-tuned on \textsc{Dpr}. We hold out $10\%$ of the \textsc{Dpr} train set ($131$ examples) to use them as the validation set. All datasets, other than \textsc{Gap}, were inspired by the Winograd Schema Challenge and come from a similar distribution. We use this model as the baseline for Scenario B.

\paragraph*{\textsc{Bert\_WikiCREM\_Dpr}.} This model is obtained by fine-tuning \textsc{Bert\_WikiCREM} on \textsc{Dpr} using the same split as for \textsc{BERT\_Dpr}. It serves as the evaluation of \textsc{WikiCREM} under Scenario~B.

\paragraph*{\textsc{Bert\_Gap\_Dpr}.} This model serves as an additional comparison to the \textsc{Bert\_WikiCREM\_Dpr} model. It is obtained by fine-tuning \textsc{Bert\_Gap} on the \textsc{Dpr} dataset.

\paragraph*{\textsc{Bert\_Gap}.} This model is obtained by fine-tuning \textsc{Bert} on the  \textsc{Gap} dataset. It serves as the baseline for Scenario C, as explained at the beginning of Section~\ref{section-experiments}.

\paragraph*{\textsc{Bert\_WikiCREM\_Gap}.} This model serves as the evaluation of \textsc{WikiCREM} for Scenario C and is obtained by fine-tuning \textsc{Bert\_WikiCREM} on \textsc{Gap}.

\paragraph*{\textsc{Bert\_all}.} This model is obtained by fine-tuning \textsc{Bert} on all the available data from the target datasets at once.
Combined \textsc{Gap}-train and \textsc{Dpr}-train data are used for training.
The model is validated on the  \textsc{Gap}-validation set and the  \textsc{WinoBias}-validation set separately.
Scores on both sets are then averaged to obtain the validation performance.
Since both training sets and both validation sets have roughly the same size, both tasks are represented equally.
%Similarly to \textsc{Bert\_Gap}, this dataset covers Scenario C.

\paragraph*{\textsc{Bert\_WikiCREM\_all}.} This model is obtained in the same way as the \textsc{Bert\_all} model, but starting from \textsc{Bert\_WikiCREM} instead. %It falls under Scenario C.

\subsection{Results}
\label{section-results}

\begin{table*}[t!]
\centering
    \begin{tabular}{@{}l@{\,}||c@{\ }c@{\ }c@{\ }c@{\ }||c@{\ }|@{\ }c@{\ }|@{\ }c@{\ }|c@{\ }}
      & \multicolumn{4}{|c||}{Transductive scenario} \\ \cline{1-8}
    & \textsc{Gap} $F_1$ & \textsc{Gap} $F_1^F$ & \textsc{Gap} $F_1^M$ & Bias $\frac{F_1^F}{F_1^M}$ & \textsc{Dpr} & \textsc{Wsc} & \textsc{Wnli} \\ \cline{1-8}
    SOTA & $72.1\%$ & $71.4\%$ & $72.8\%$ & $0.98$ & $76.4\%$ & $\underline{\mathbf{72.5}\%}$  & $\underline{\mathbf{74.7}\%}$\\ \hline \hline
    \textsc{Bert} & $50.0\%$ & $47.2\%$ & $52.7\%$ & $0.90$ & $59.8\%$ & $61.9\%$ & $65.8\%$ & \multirow{3}{2.5em}{no train data}\\ 
    \textsc{Bert\_WikiRand} & $55.1\%$ & $51.8\%$ & $58.2\%$ & $0.89$ & $59.2\%$ & $59.3\%$ & $65.8\%$\\ 
    \textsc{Bert\_WikiCREM} & $\mathbf{59.0}\%$ & $\mathbf{57.5}\%$ & $\mathbf{60.5}\%$ & $\mathbf{0.95}$ & $\mathbf{67.4}\%$ & $\mathbf{63.4}\%$  & $\mathbf{67.1}\%$\\ \hhline{-||----||====} 
    \textsc{Bert\_Gap} & $75.2\%$ & $75.1\%$ & $75.3\%$ & $\underline{\mathbf{1.00}}$ & $66.8\%$ & $63.0\%$ & $68.5\%$ & \multirow{7}{2.5em}{existing train data}\\
    \textsc{Bert\_WikiCREM\_Gap} & $\mathbf{77.4}\%$ & $\mathbf{78.4}\%$ & $\mathbf{76.4}\%$ & $1.03$  & $\mathbf{71.1}\%$ & $\mathbf{64.1}\%$ & $\mathbf{70.5}\%$\\ \cline{1-8}
    \textsc{Bert\_Dpr} & $60.9\%$ & $61.3\%$ & $60.6\%$ & $1.01$ & $\mathbf{83.3}\%$ & $67.0\%$ & $71.9\%$\\
    \textsc{Bert\_Gap\_Dpr} & $\mathbf{70.0}\%$ & $\mathbf{70.4}\%$ & $\mathbf{69.5}\%$ & $1.01$ & $79.4\%$ & $65.6\%$ & $72.6\%$\\
    \textsc{Bert\_WikiCREM\_Dpr} & $64.2\%$ & $64.2\%$ & $64.1\%$ & $\underline{\mathbf{1.00}}$  & $80.0\%$ & $\mathbf{71.8}\%$ & $\underline{\mathbf{74.7}\%}$\\ \cline{1-8}
    \textsc{Bert\_all} & $76.0\%$ & $77.4\%$ & $74.7\%$ & $\mathbf{1.04}$ & $80.1\%$ & $\mathbf{70.0}\%$ & $74.0\%$\\
    \textsc{Bert\_WikiCREM\_all} & $\underline{\mathbf{78.0}\%}$ & $\underline{\mathbf{79.4}\%}$ & $\underline{\mathbf{76.7}\%}$ & $\mathbf{1.04}$ & $\underline{\mathbf{84.8}\%}$ & $\mathbf{70.0}\%$ & $\underline{\mathbf{74.7}\%}$\\ \cline{1-8}
    \end{tabular}
    \vspace{2mm}\\
    \begin{tabular}{@{}l@{\,}|c@{\ }c@{\ }c@{\ }c@{\ }|@{\ }c@{\ }|@{\ }c@{\ }|c@{\ }}
    \cline{1-7}
    & \textsc{WB} T1-a & \textsc{WB} T1-p & \textsc{WB} T2-a & \textsc{WB} T2-p & \textsc{WinoGender} & \textsc{Pdp} \\  \cline{1-7}
    SOTA & $60.6\%$ & $74.9\%$ & $78.0\%$ & $88.6\%$ & $50.9\%$ & $74.0\%$\\ \hline \hline
    \textsc{Bert} & $61.3\%$ & $60.3\%$ & $76.2\%$ & $75.8\%$ & $59.2\%$ & $71.7\%$ & \multirow{3}{2.5em}{no train data}\\
    \textsc{Bert\_WikiRand} & $53.5\%$ & $52.5\%$ & $64.6\%$ & $65.2\%$ & $57.9\%$ & $73.3\%$ \\
    \textsc{Bert\_WikiCREM} & $\mathbf{65.2}\%$ & $\mathbf{64.9}\%$ & $\mathbf{95.7}\%$ & $\mathbf{94.9}\%$ & $\mathbf{66.7}\%$  & $\mathbf{76.7}\%$\\ \hline \hline
    \textsc{Bert\_Gap} & $64.6\%$ & $63.8\%$ & $88.1\%$ & $87.9\%$ & $67.5\%$ & $\mathbf{85.0}\%$ & \multirow{7}{2.5em}{existing train data}\\
    \textsc{Bert\_WikiCREM\_Gap} & $\mathbf{71.2}\%$ & $\mathbf{70.5}\%$ & $\mathbf{97.2}\%$ & $\mathbf{98.2}\%$ & $\mathbf{75.4}\%$ & $83.3\%$\\  \cline{1-7}
    \textsc{Bert\_Dpr} & $\underline{\mathbf{78.0}\%}$ & $\underline{\mathbf{78.2}\%}$ & $85.6\%$ & $86.4\%$ & $79.2\%$ & $81.7\%$\\
    \textsc{Bert\_Gap\_Dpr} & $77.8\%$ & $76.5\%$ & $\mathbf{89.6}\%$ & $\mathbf{89.1}\%$ & $75.8\%$ & $\underline{\mathbf{86.7}\%}$\\
    \textsc{Bert\_WikiCREM\_Dpr} & $76.0\%$ & $76.3\%$ & $81.3\%$ & $80.3\%$ & $\underline{\mathbf{82.1}\%}$ & $76.7\%$\\  \cline{1-7}
    \textsc{Bert\_all} & $\mathbf{77.8}\%$ & $\mathbf{77.2}\%$ & $94.7\%$ & $94.9\%$ & $\mathbf{78.8}\%$ & $81.7\%$ \\
    \textsc{Bert\_WikiCREM\_all} & $76.8\%$ & $75.8\%$ & $\underline{\mathbf{98.7}\%}$ & $\underline{\mathbf{99.0}\%}$ & $76.7\%$ & $\underline{\mathbf{86.7}\%}$ \\  \cline{1-7}
    \end{tabular}
    
    \caption{Evaluation of trained models on all test sets. \textsc{Gap} and \textsc{WinoBias} (abbreviated \textsc{WB}) are additionally split into subsets, as introduced in Section~\ref{section-datasets}. 
    Double lines in the table separate results from three different scenarios: when no training data is available, when additional training data exists, and the transductive scenario.
    The table is further split into sections separated with single horizontal lines. Each section contains a model that has been trained on \textsc{WikiCREM} and models that have not been. The best result in each section is in bold. The best overall result is underlined. Scores on \textsc{Gap} are measured as $F_1$-scores, while the performance on other datasets is given in accuracy. The source of each SOTA is listed in Section~\ref{section-datasets}.\vspace*{-3ex}}
    \label{table-results}
\end{table*}

The results of the evaluation of the models on the test sets are shown in Table \ref{table-results}.
We notice that additional training on \textsc{WikiCREM} consistently improves the performance of the models in all scenarios and on most tests.
Due to the small size of some test sets, some of the results are subject to deviation.
This especially applies to  \textsc{Pdp} ($60$ test samples) and \textsc{Wnli} ($145$ test samples).

We observe that \textsc{Bert\_WikiRand} generally performs worse than \textsc{Bert}, with \textsc{Gap} and \textsc{Pdp} being notable exceptions.
This shows that \textsc{Bert} is a strong baseline and that improved performance of \textsc{Bert\_WikiCREM} is not a consequence of training on shorter sentences or with different loss function.
\textsc{Bert\_WikiCREM} consistently out\-performs both baselines on all tests, showing that \textsc{WikiCREM} can be used as a standalone dataset.

We observe that training on the data from the target distribution improves the performance the most.
Models trained on \textsc{Gap}-train usually show more than a $20\%$ increase in their $F_1$-score on \textsc{Gap}-test.
Still, \textsc{Bert\_WikiCREM\_Gap} shows a consistent improvement over \textsc{Bert\_Gap} on all subsets of the \textsc{Gap} test set.
%Additionally, it exhibits the least bias (the closer the score to $1$, the better).
This confirms that \textsc{WikiCREM} works not just as a standalone dataset, but also as an additional pre-training in the transductive scenario.

Similarly, \textsc{Bert\_WikiCREM\_Dpr} outperforms \textsc{Bert\_Dpr} on the majority of tasks, showing the applicability of \textsc{WikiCREM} to the scenario where additional training data is available.
However, good results of \textsc{Bert\_Gap\_Dpr} show that additional training on a manually constructed dataset, such as \textsc{Gap}, can yield similar results as additional training on \textsc{WikiCREM}.
The reason behind this difference is the impact of the data distribution.
\textsc{Gap}, \textsc{Dpr}, and \textsc{WikiCREM} contain data that follows different distributions which strongly impacts the trained models.
This can be seen when we fine-tune \textsc{Bert\_Gap} on \textsc{Dpr} to obtain \textsc{Bert\_Gap\_Dpr}, as the model's performance on \textsc{Gap}-test drops by $8.2\%$.
\textsc{WikiCREM}'s data distribution strongly differs from the test sets' as described in Section~\ref{section-dataset}.%, $37$ in $100$ examples do not form a natural sentence if a pronoun is put in the place of the \texttt{[MASK]} token.
%This difference is likely the cause of the worse performance of \textsc{Bert\_WikiCREM\_Dpr} in comparison to \textsc{Bert\_Gap\_Dpr}.

However, the best results are achieved when all available data is combined, as shown by the models \textsc{Bert\_all} and \textsc{Bert\_WikiCREM\_all}.
\textsc{Bert\_WikiCREM\_all} achieves the highest performance on \textsc{Gap}, \textsc{Dpr}, \textsc{Wnli}, and \textsc{WinoBias} among the models, and sets the new state-of-the-art result on \textsc{Gap}, \textsc{Dpr}, and \textsc{WinoBias}.
The new state-of-the-art result on the \textsc{WinoGender} dataset is achieved by the \textsc{Bert\_WikiCREM\_Dpr} model, while \textsc{Bert\_WikiCREM\_all} and \textsc{Bert\_Gap\_Dpr} set the new state-of-the-art on the \textsc{Pdp} dataset.

\section{Conclusions and Future Work}
In this work, we introduced \textsc{WikiCREM}, a large dataset of training instances for pronoun resolution.
%It contains $2.4$M instances of real-world pronoun disambiguation sentences, collected from Wikipedia. 
We use our dataset to fine-tune the \textsc{Bert} language model. Our results match or outperform state-of-the-art models on $6$ out of $7$ evaluated datasets. 

The employed data-generating procedure can be further applied to other large sources of text to generate more training sets for pronoun resolution. In addition, both variety and size of the generated datasets can be increased if we do not restrict ourselves to personal names.
We hope that the community will make use of our released \textsc{WikiCREM} dataset to further improve the pronoun resolution task.

\section*{Acknowledgments}
This work was supported  by the Alan Turing Institute under the UK EPSRC grant EP/N510129/1, by the EPSRC grant EP/R013667/1, by the EPSRC studentship OUCS/EPSRC-NPIF/VK/ 1123106, by the JP Morgan PhD Fellowship 2019-2020, and 
by an EPSRC Vacation Bursary. 
We also acknowledge the use of the EPSRC-funded Tier 2 facility JADE (EP/P020275/1).
\bibliographystyle{acl_natbib}
\bibliography{refs}
\appendix
\section{Annotated \textsc{WikiCREM} Examples}
\label{appendix}
The appendix contains the $100$ manually annotated examples.

\begin{enumerate}
\item Throughout training camp , Jackson competed to be the Bengals ' third cornerback on the depth chart against Darqueze Dennard . On August 2 , 2016 , it was announced that \texttt{[MASK]} had suffered a torn pectoral muscle and would have to undergo surgery . \\ 
\textbf{ ambiguous }\\ Pronoun in place of \texttt{[MASK]}?: \textbf{ No }\\ 
Annotator's answer: \textbf{ N/A }\\ Correct?: \textbf{ N/A } 
\item The Ark " consisted of a giant rowboat with a small engine which Beek used as his first ferry vessel . " The \texttt{[MASK]} " carried oars in the event of engine failure . \\ 
\textbf{ not ambiguous }\\ Pronoun in place of \texttt{[MASK]}?: \textbf{ no }\\ 
Annotator's answer: \textbf{ Ark }\\ Correct?: \textbf{ yes } 
\item However , John was able to gain the lost estates by a marriage to Joanna of Urgell , granddaughter of Peter IV of Aragon . \texttt{[MASK]} fought with Aragon against Castile , but helped his brother Peter , Cardinal of Foix and Arles , to crush insurgents from Aragon . \\ 
\textbf{ not ambiguous }\\ Pronoun in place of \texttt{[MASK]}?: \textbf{ yes }\\ 
Annotator's answer: \textbf{ John }\\ Correct?: \textbf{ yes } 
\item Ultravox had gone on to greater success with Midge Ure fronting the band , but when Simon left the band in 1988 , Billy Currie formed a new band which later included \texttt{[MASK]} . \\ 
\textbf{ not ambiguous }\\ Pronoun in place of \texttt{[MASK]}?: \textbf{ no }\\ 
Annotator's answer: \textbf{ Midge Ure }\\ Correct?: \textbf{ no } 
\item The poem describes the poet 's idyllic family life with his own three daughters , Alice , Edith , and Anne Allegra : " grave \texttt{[MASK]} , and laughing Allegra , and Edith with golden hair . " \\ 
\textbf{ not ambiguous }\\ Pronoun in place of \texttt{[MASK]}?: \textbf{ no   }\\ 
Annotator's answer: \textbf{ Alice }\\ Correct?: \textbf{ yes } 
\item Koch and Eide searched Cho 's belongings and found a pocket knife , but they did not find any items that they deemed threatening . \texttt{[MASK]} also described a telephone call that he received from Cho during the Thanksgiving holiday break from school . \\ 
\textbf{ ambiguous }\\ Pronoun in place of \texttt{[MASK]}?: \textbf{ no }\\ 
Annotator's answer: \textbf{ N/A }\\ Correct?: \textbf{ N/A } 
\item As Rajveer was able to successfully lead the escape of them both , Harleen now entrusts Rajveer , subsequently falling in love with him . When Rajveer goes out of his hotel with \texttt{[MASK]} , he sees that they are wanted by Interpol . \\ 
\textbf{ not ambiguous }\\ Pronoun in place of \texttt{[MASK]}?: \textbf{ yes }\\ 
Annotator's answer: \textbf{ Harleen }\\ Correct?: \textbf{ yes } 
\item Elmas , Su Masu in Sardinian language , is a " comune " of the Metropolitan City of Cagliari in the Italian region of Sardinia , located about northwest of Cagliari . Until 1989 \texttt{[MASK]} was a district of Cagliari . \\ 
\textbf{ not ambiguous }\\ Pronoun in place of \texttt{[MASK]}?: \textbf{ yes }\\ 
Annotator's answer: \textbf{ Elmas }\\ Correct?: \textbf{ yes } 
\item Later in the year , Li Keyong did send Li Sizhao and Zhou to capture Xi and Ci Prefectures , which had become under Zhu 's control when \texttt{[MASK]} conquered Huguo earlier in 901 . \\ 
\textbf{ not ambiguous }\\ Pronoun in place of \texttt{[MASK]}?: \textbf{ yes }\\ 
Annotator's answer: \textbf{ Zhu }\\ Correct?: \textbf{ yes } 
\item Elisha told Hazael to tell Hadadezer that he would recover , and he revealed to Hazael that the king would recover but would die of other means . The day after he returned to Hadadezer in Damascus , \texttt{[MASK]} suffocated him and seized power himself . \\ 
\textbf{ not ambiguous }\\ Pronoun in place of \texttt{[MASK]}?: \textbf{ yes }\\ 
Annotator's answer: \textbf{ Hazael }\\ Correct?: \textbf{ yes } 
\item In 1946 , Hill and Knowlton dissolved their partnership , and Knowlton took over the direction of Hill \& Knowlton Cleveland , which closed shortly after Knowlton's retirement in 1962 . \texttt{[MASK]} also maintained an interest in music . \\ 
\textbf{ ambiguous }\\ Pronoun in place of \texttt{[MASK]}?: \textbf{ no }\\ 
Annotator's answer: \textbf{ N/A }\\ Correct?: \textbf{ N/A } 
\item First , God is revealed with Law , and secondly , God is revealed as Person . \texttt{[MASK]} 's anger at Moses for not speaking to the rock on the second occasion , highlights that this is not the spiritual picture He wanted portrayed . \\ 
\textbf{ not ambiguous }\\ Pronoun in place of \texttt{[MASK]}?: \textbf{ yes }\\ 
Annotator's answer: \textbf{ God }\\ Correct?: \textbf{ yes } 
\item He now said he had seen Acreman follow Cheryl Fergeson up a staircase leading to the auditorium and then heard her scream , " No " and " Don 't . " Later that day , \texttt{[MASK]} warned Sessum not to tell anyone what he had seen . \\ 
\textbf{ not ambiguous }\\ Pronoun in place of \texttt{[MASK]}?: \textbf{ no }\\ 
Annotator's answer: \textbf{ Acreman }\\ Correct?: \textbf{ yes } 
\item Meidi finally figures it out , but does not reveal to Qi Yue and Ah Meng until \texttt{[MASK]} confesses . \\ 
\textbf{ ambiguous }\\ Pronoun in place of \texttt{[MASK]}?: \textbf{ no }\\ 
Annotator's answer: \textbf{ N/A }\\ Correct?: \textbf{ N/A } 
\item Brett went back to Leary , expecting to be turned down again , but this time , Leary gave Brett the aircraft he wanted . " Perhaps " , \texttt{[MASK]} speculated , " Leary had heard from Washington " . \\ 
\textbf{ not ambiguous }\\ Pronoun in place of \texttt{[MASK]}?: \textbf{ yes }\\ 
Annotator's answer: \textbf{ Brett }\\ Correct?: \textbf{ yes } 
\item Maurice White spoke to Stepney on the morning of May 17 , 1976 , but later that day , Earth , Wind \& Fire keyboardist Larry Dunn received a phone call , informing him that \texttt{[MASK]} had died of a heart attack . \\ 
\textbf{ not ambiguous }\\ Pronoun in place of \texttt{[MASK]}?: \textbf{ no }\\ 
Annotator's answer: \textbf{ Stepney }\\ Correct?: \textbf{ yes } 
\item Li Yi sought aid from Gao , who personally led two thousand cavalry soldiers to aid Li Yi , causing Dou to withdraw . Gao thereafter sought to submit to Tang , through \texttt{[MASK]} . \\ 
\textbf{ not ambiguous }\\ Pronoun in place of \texttt{[MASK]}?: \textbf{ no }\\ 
Annotator's answer: \textbf{ Li Yi }\\ Correct?: \textbf{ yes } 
\item At Cambridge , Rose studied under Hubert Middleton and Edward Joseph Dent from 1935 to 1939 . \texttt{[MASK]} started his academic career at The Queen 's College , Oxford . \\ 
\textbf{ not ambiguous }\\ Pronoun in place of \texttt{[MASK]}?: \textbf{ yes }\\ 
Annotator's answer: \textbf{ Rose }\\ Correct?: \textbf{ yes } 
\item Brady and Bolger leave with Rothbaum , and Rothbaum demands the money Brady owes him . When Rothbaum threatens to kill them if they don 't pay up , Bolger shoots Rothbaum 's thugs , and Brady stabs \texttt{[MASK]} , killing him . \\ 
\textbf{ not ambiguous }\\ Pronoun in place of \texttt{[MASK]}?: \textbf{ no }\\ 
Annotator's answer: \textbf{ Rothbaum }\\ Correct?: \textbf{ yes } 
\item In the reception room , a boy named Billy won 't stop staring at Don . \texttt{[MASK]} is drawing a picture and then rips it out of his book and hands it to Billy , getting up and leaving . \\ 
\textbf{ not ambiguous }\\ Pronoun in place of \texttt{[MASK]}?: \textbf{ yes }\\ 
Annotator's answer: \textbf{ Don }\\ Correct?: \textbf{ yes } 
\item Hasan and Stein agree that Harsa became king in 1089 . Utkarsa was disliked and soon deposed , with a half - brother called Vijayamalla supporting \texttt{[MASK]} and being at the forefront of the rebellion against the king . \\ 
\textbf{ ambiguous }\\ Pronoun in place of \texttt{[MASK]}?: \textbf{ no }\\ 
Annotator's answer: \textbf{ N/A }\\ Correct?: \textbf{ N/A } 
\item Yet this did not prevent Leisegang from reasserting that Aristotle 's own pattern of thinking was incompatible with a proper understanding of Plato . " Therein Cherniss believed Jaeger to be contrary to \texttt{[MASK]} , and Leisegang was unsympathetic to compatibility between Plato and Aristotle in both and above . \\ 
\textbf{ not ambiguous }\\ Pronoun in place of \texttt{[MASK]}?: \textbf{ no }\\ 
Annotator's answer: \textbf{ Leisegang }\\ Correct?: \textbf{ yes } 
\item Aska later decides not to rule Cephiro because Fuu told her that the Pillar can think only of Cephiro , but since Lady Aska loves the people of Fahren , she cannot complete the task of Pillar in \texttt{[MASK]} . \\ 
\textbf{ not ambiguous }\\ Pronoun in place of \texttt{[MASK]}?: \textbf{ yes }\\ 
Annotator's answer: \textbf{ Cephiro }\\ Correct?: \textbf{ yes } 
\item As the weeks wore on , it became evident that Nick could be Lujack 's twin brother , hence , Alex 's son . \texttt{[MASK]} soon became obsessed with Nick and Mindy warned him that Alex wouldn 't give up until she got him . \\ 
\textbf{ not ambiguous }\\ Pronoun in place of \texttt{[MASK]}?: \textbf{ yes }\\ 
Annotator's answer: \textbf{ Alex }\\ Correct?: \textbf{ yes } 
\item Due to Jijii 's psychological manipulation , Ichi believes that Kaneko is his brother and confronts him . Kaneko shoots the side of \texttt{[MASK]} 's leg , causing Ichi to slit Kaneko 's throat in front of Takeshi . \\ 
\textbf{ not ambiguous }\\ Pronoun in place of \texttt{[MASK]}?: \textbf{ yes }\\ 
Annotator's answer: \textbf{ Ichi }\\ Correct?: \textbf{ yes } 
\item Frank , Jump and other members of the gang go to Clay 's social club , where Frank tells Clay that he wants a percentage of all Clay 's profits . When Clay insults him , \texttt{[MASK]} shoots the Mafioso . \\ 
\textbf{ ambiguous }\\ Pronoun in place of \texttt{[MASK]}?: \textbf{ no  }\\ 
Annotator's answer: \textbf{ N/A }\\ Correct?: \textbf{ N/A } 
\item The spoils were to be divided between Shivaji , Kootab Shah and Bijapur . With the agreement concluded and with \texttt{[MASK]} giving him money , horses and artillery , Sivajee set out in March 1677 for his invasions via Kurnool , Cuddapah and Madras . \\ 
\textbf{ ambiguous }\\ Pronoun in place of \texttt{[MASK]}?: \textbf{ no  }\\ 
Annotator's answer: \textbf{ N/A }\\ Correct?: \textbf{ N/A } 
\item Billingsley 's response was a gift—bow ties for Ace . \texttt{[MASK]} 's reply was to ask Billingsley for some matching socks so he would be well - dressed when he was refused admittance again . \\ 
\textbf{ not ambiguous }\\ Pronoun in place of \texttt{[MASK]}?: \textbf{ yes }\\ 
Annotator's answer: \textbf{ Ace }\\ Correct?: \textbf{ yes } 
\item In 890 , when Zhu asked Luo Hongxin the military governor of Weibo Circuit for permission to go through Luo 's territory to attack Hedong , Luo refused . \texttt{[MASK]} reacted by sending Ding , Ge , Pang Shigu , and Huo Cun to attack Weibo . \\ 
\textbf{ not ambiguous }\\ Pronoun in place of \texttt{[MASK]}?: \textbf{ yes }\\ 
Annotator's answer: \textbf{ Zhu }\\ Correct?: \textbf{ yes } 
\item However , it is recorded that Lewis was born in 1381 and sent to the school at Oxford at age 10 ; it is also known that Chaucer 's " Treatise on the Astrolabe " was written for \texttt{[MASK]} . \\ 
\textbf{ not ambiguous }\\ Pronoun in place of \texttt{[MASK]}?: \textbf{ yes }\\ 
Annotator's answer: \textbf{ Lewis }\\ Correct?: \textbf{ yes } 
\item Claiborne and the other survivors are rescued , thanks to quick action by Taylor and Harold . At the interment , \texttt{[MASK]} begs Claiborne to take him and Harold on his expedition to K2 , the second highest peak in the world . \\ 
\textbf{ not ambiguous }\\ Pronoun in place of \texttt{[MASK]}?: \textbf{ yes }\\ 
Annotator's answer: \textbf{ Taylor }\\ Correct?: \textbf{ yes } 
\item That year , LONGi signed a contract with Yingli to cooperate on monocrystalline products . In early 2016 , \texttt{[MASK]} signed a \$1 . \\ 
\textbf{ ambiguous }\\ Pronoun in place of \texttt{[MASK]}?: \textbf{ no  }\\ 
Annotator's answer: \textbf{ N/A }\\ Correct?: \textbf{ N/A } 
\item Katie is taken to Children 's Hospital , and Louise and Wes find themselves being arrested for " what authorities are calling the worst case of child abuse they 've ever seen " . Shortly before his trial begins , \texttt{[MASK]} kills himself . \\ 
\textbf{ not ambiguous }\\ Pronoun in place of \texttt{[MASK]}?: \textbf{ yes }\\ 
Annotator's answer: \textbf{ Wes }\\ Correct?: \textbf{ yes } 
\item When it becomes clear that Warren has shifted his interest from Marjorie to Bernice , Marjorie sets about humiliating Bernice by tricking her into going through with bobbing her hair . When \texttt{[MASK]} comes out of the barbershop with the new hairdo , her hair is flat and strange . \\ 
\textbf{ not ambiguous }\\ Pronoun in place of \texttt{[MASK]}?: \textbf{ yes }\\ 
Annotator's answer: \textbf{ Bernice }\\ Correct?: \textbf{ yes } 
\item Rebello defeated David Cho via Unanimous Decision at PXC 26 - Meanest Game Face on August 20 , 2011 . \texttt{[MASK]} made his WEC debut at WEC 39 , losing to Kenji Osawa via split decision . \\ 
\textbf{ not ambiguous }\\ Pronoun in place of \texttt{[MASK]}?: \textbf{ yes }\\ 
Annotator's answer: \textbf{ Rebello }\\ Correct?: \textbf{ yes } 
\item Karen overhears Bernadette talking to Keanu , who \texttt{[MASK]} thinks may be the baby 's father . \\ 
\textbf{ not ambiguous }\\ Pronoun in place of \texttt{[MASK]}?: \textbf{ yes }\\ 
Annotator's answer: \textbf{ Karen }\\ Correct?: \textbf{ yes } 
\item Murray Abraham , Daphne Rubin - Vega , Henry Winkler , Kathryn Boule and Judy Kuhn also got their start with Theatreworks . \texttt{[MASK]} has won many awards in its long history . \\ 
\textbf{ not ambiguous }\\ Pronoun in place of \texttt{[MASK]}?: \textbf{ yes }\\ 
Annotator's answer: \textbf{ Theatreworks }\\ Correct?: \textbf{ yes } 
\item In Jacmel , three weeks prior to his reunion with Marie , Paul spends time with his lover and fiancée Natasha . Natasha harbors feelings of mistrust for \texttt{[MASK]} , who left for New York after the earthquake , and spent three years there without having ever contacted her . \\ 
\textbf{ not ambiguous }\\ Pronoun in place of \texttt{[MASK]}?: \textbf{ yes }\\ 
Annotator's answer: \textbf{ Marie }\\ Correct?: \textbf{ no } 
\item After the match , Deuce 'n Domino attacked Snuka and Slaughter until Tony Garea and Rick Martel came into the ring to assist Snuka and \texttt{[MASK]} . \\ 
\textbf{ not ambiguous }\\ Pronoun in place of \texttt{[MASK]}?: \textbf{ yes }\\ 
Annotator's answer: \textbf{ Slaughter }\\ Correct?: \textbf{ yes } 
\item Later , Jude and Noah realize that they will be working together , as \texttt{[MASK]} is a new sideline reporter assigned to the Devils . \\ 
\textbf{ ambiguous }\\ Pronoun in place of \texttt{[MASK]}?: \textbf{ no  }\\ 
Annotator's answer: \textbf{ N/A }\\ Correct?: \textbf{ N/A } 
\item As such , Michael , Madeline , Sam , Fiona , and Jesse are all hell - bent on exacting revenge for Nate 's murder . Eventually , Michael , with his former mentor Tom Card helping him , tracks \texttt{[MASK]} 's killer , Tyler Grey , to Panama . \\ 
\textbf{ not ambiguous }\\ Pronoun in place of \texttt{[MASK]}?: \textbf{ yes }\\ 
Annotator's answer: \textbf{ Nate }\\ Correct?: \textbf{ yes } 
\item Eileen later follows Des to Erinsborough to check up on him and she takes an instant dislike to \texttt{[MASK]} ' housemate , Daphne Lawrence . \\ 
\textbf{ not ambiguous }\\ Pronoun in place of \texttt{[MASK]}?: \textbf{ yes }\\ 
Annotator's answer: \textbf{ Des }\\ Correct?: \textbf{ yes } 
\item Although Armstrong was a third party not in privity with Leyland , and a stranger to the car purchase transaction , nonetheless Armstrong was permitted to rely on the non - derogation rights of the car owners relative to \texttt{[MASK]} . \\ 
\textbf{ not ambiguous }\\ Pronoun in place of \texttt{[MASK]}?: \textbf{ yes }\\ 
Annotator's answer: \textbf{ Leyland }\\ Correct?: \textbf{ yes } 
\item Its theological center and the Fatima Masumeh Shrine are prominent features of Qom . Another very popular religious site of pilgrimage formerly outside the city of \texttt{[MASK]} but now more of a suburb is called Jamkaran . \\ 
\textbf{ not ambiguous }\\ Pronoun in place of \texttt{[MASK]}?: \textbf{ yes }\\ 
Annotator's answer: \textbf{ Qom }\\ Correct?: \textbf{ yes } 
\item Schult was married to Elke Koska for 25 years , who Schult considers his muse - she was also his manager , now in cooperation with Anna Zlotovskaya , the Russian classical violinist , \texttt{[MASK]} married in 2010 . \\ 
\textbf{ not ambiguous }\\ Pronoun in place of \texttt{[MASK]}?: \textbf{ yes }\\ 
Annotator's answer: \textbf{ Schult }\\ Correct?: \textbf{ yes } 
\item Gordon wakes up and successfully escapes from Nygma . Shaking off \texttt{[MASK]} 's pursuit , Gordon reaches Bruce and Selina 's hideout and collapses . \\ 
\textbf{ not ambiguous }\\ Pronoun in place of \texttt{[MASK]}?: \textbf{ yes }\\ 
Annotator's answer: \textbf{ Nygma }\\ Correct?: \textbf{ yes } 
\item During World War II , Hill , as well as Lewis , filed for conscientious objector status . After the war , \texttt{[MASK]} , Hill and a small group of former conscientious objectors created the Pacifica Foundation in Pacifica , California . \\ 
\textbf{ not ambiguous }\\ Pronoun in place of \texttt{[MASK]}?: \textbf{ yes }\\ 
Annotator's answer: \textbf{ Lewis }\\ Correct?: \textbf{ yes } 
\item Lana tries to intervene but is punched in the stomach by Dino . Luca lunges at Dino but \texttt{[MASK]} pushes him to the ground . \\ 
\textbf{ not ambiguous }\\ Pronoun in place of \texttt{[MASK]}?: \textbf{ yes }\\ 
Annotator's answer: \textbf{ Dino }\\ Correct?: \textbf{ yes } 
\item On October 28 , 2010 , Facebook banned Rapleaf from scraping data on Facebook , and \texttt{[MASK]} said it would delete the Facebook IDs it had collected . \\ 
\textbf{ not ambiguous }\\ Pronoun in place of \texttt{[MASK]}?: \textbf{ yes }\\ 
Annotator's answer: \textbf{ Rapleaf }\\ Correct?: \textbf{ yes } 
\item After arriving at the camp , Charlie apologizes to Claire , but Claire tells him to leave her and her son alone . \texttt{[MASK]} then goes into the jungle , and opens a hiding place where he is keeping Virgin Mary statues to put the one Eko gave him . \\ 
\textbf{ not ambiguous }\\ Pronoun in place of \texttt{[MASK]}?: \textbf{ yes }\\ 
Annotator's answer: \textbf{ Charlie }\\ Correct?: \textbf{ yes } 
\item When Sylvie belittles Babe , she leaves Sylvie by the canal in the rain , although Sylvie is found and Shirley realises that Babe left \texttt{[MASK]} to die and disowns her . \\ 
\textbf{ not ambiguous }\\ Pronoun in place of \texttt{[MASK]}?: \textbf{ yes }\\ 
Annotator's answer: \textbf{ Sylvie }\\ Correct?: \textbf{ yes } 
\item Gregory , as an infant , drowned in a bathtub when Kay became distracted from a call from Sam . \texttt{[MASK]} and Kay ended up divorcing . \\ 
\textbf{ not ambiguous }\\ Pronoun in place of \texttt{[MASK]}?: \textbf{ yes }\\ 
Annotator's answer: \textbf{ Sam }\\ Correct?: \textbf{ yes } 
\item Stefanie in Rio is a 1960 West German romantic comedy film directed by Curtis Bernhardt and starring Carlos Thompson , Sabine Sinjen and Andréa Parisy . It is a sequel to the 1958 film " \texttt{[MASK]} " . \\ 
\textbf{ ambiguous }\\ Pronoun in place of \texttt{[MASK]}?: \textbf{ no }\\ 
Annotator's answer: \textbf{ N/A }\\ Correct?: \textbf{ N/A } 
\item Kent Ling and his team of assassins are then forced to rescue Ling Hung , but it involved them and Ling Hung having to be in a very deadly gun battle against Kam Tin 's henchmen and unfortunately \texttt{[MASK]} 's team are all killed in the process . \\ 
\textbf{ not ambiguous }\\ Pronoun in place of \texttt{[MASK]}?: \textbf{ yes }\\ 
Annotator's answer: \textbf{ Kent Ling }\\ Correct?: \textbf{ yes } 
\item On the weekend of January 14 , 2017 , Walker was planning to compete at the 2016 Montana ProRodeo Circuit Finals in Great Falls . \texttt{[MASK]} was in 2nd place in the circuit standings with \$14 , 351 so far . \\ 
\textbf{ not ambiguous }\\ Pronoun in place of \texttt{[MASK]}?: \textbf{ yes }\\ 
Annotator's answer: \textbf{ Walker }\\ Correct?: \textbf{ yes } 
\item Samantha , Jennifer , Billy , Taylor and Coop leave by the end of the season . In the season finale , \texttt{[MASK]} gives birth to Michael 's son and agrees to share motherhood with the returning Jane . \\ 
\textbf{ ambiguous }\\ Pronoun in place of \texttt{[MASK]}?: \textbf{ no }\\ 
Annotator's answer: \textbf{ N/A }\\ Correct?: \textbf{ N/A } 
\item Appears in " " Bucky was a worker who encountered Michael Myers as he wandered around an electrical power plant . \texttt{[MASK]} told Michael that he was not permitted on the grounds . \\ 
\textbf{ not ambiguous }\\ Pronoun in place of \texttt{[MASK]}?: \textbf{ yes }\\ 
Annotator's answer: \textbf{ Bucky }\\ Correct?: \textbf{ yes } 
\item In August of the same year , Laura came to Coronation Street to tell Alan and Elsie Tanner that she was getting re - married and wanted to drop \texttt{[MASK]} 's loan - although Elsie refused . \\ 
\textbf{ not ambiguous }\\ Pronoun in place of \texttt{[MASK]}?: \textbf{ yes }\\ 
Annotator's answer: \textbf{ Elsie }\\ Correct?: \textbf{ no } 
\item When Spike lands , Jerry sticks out his tongue and throws Spike 's lips over his own face , provoking \texttt{[MASK]} to chase him around the corner . \\ 
\textbf{ not ambiguous }\\ Pronoun in place of \texttt{[MASK]}?: \textbf{ yes }\\ 
Annotator's answer: \textbf{ Spike }\\ Correct?: \textbf{ yes } 
\item A year later she met Friedrich Schiller and played Luise Miller in his first performance of Kabale und Liebe . Sophie Albrecht and \texttt{[MASK]} had similar interests and became close friends . \\ 
\textbf{ not ambiguous }\\ Pronoun in place of \texttt{[MASK]}?: \textbf{ yes }\\ 
Annotator's answer: \textbf{ Friedrich Schiller }\\ Correct?: \textbf{ yes } 
\item Walcott lost the count as Ali circled around a floored Liston and \texttt{[MASK]} tried to get him back to a neutral corner . \\ 
\textbf{ not ambiguous }\\ Pronoun in place of \texttt{[MASK]}?: \textbf{ yes }\\ 
Annotator's answer: \textbf{ Walcott }\\ Correct?: \textbf{ yes } 
\item At the World Matchplay , Whitlock recorded wins over Kevin Painter , Raymond van Barneveld and James Wade to reach the semi - finals of the event for the second time , with \texttt{[MASK]} stating he was playing his best darts in five years . \\ 
\textbf{ ambiguous }\\ Pronoun in place of \texttt{[MASK]}?: \textbf{ no }\\ 
Annotator's answer: \textbf{ N/A }\\ Correct?: \textbf{ N/A } 
\item In the long period that Lars Semb was manager at Moss Jernverk he traveled almost yearly to the mining areas and he subsequently stayed with the local agents . \texttt{[MASK]} was totally dependent on charcoal that the surrounding farmers produced . \\ 
\textbf{ not ambiguous }\\ Pronoun in place of \texttt{[MASK]}?: \textbf{ yes }\\ 
Annotator's answer: \textbf{ Moss Jernverk }\\ Correct?: \textbf{ yes } 
\item Armstrong felt impressed with the style of Hansen 's work . In June 2002 , Armstrong and \texttt{[MASK]} signed a formal agreement . \\ 
\textbf{ not ambiguous }\\ Pronoun in place of \texttt{[MASK]}?: \textbf{ yes }\\ 
Annotator's answer: \textbf{ Hansen }\\ Correct?: \textbf{ yes } 
\item In 1235 and 1239 the da Camino managed to obtain the rule in Treviso , but the second time they were betrayed by Alberico da Romano , who expelled the Guelphs from the city . However , with Gherardo III da Camino the \texttt{[MASK]} regained prominence . \\ 
\textbf{ not ambiguous }\\ Pronoun in place of \texttt{[MASK]}?: \textbf{ yes }\\ 
Annotator's answer: \textbf{ Guelphs }\\ Correct?: \textbf{ yes } 
\item The small forward Shamell Stallworth made a three - pointer with the clock reset already and gave the victory to Pinheiros . This game was extremely important because \texttt{[MASK]} because Pinheiros finished the regular season in front of Flamengo precisely by direct confrontation . \\ 
\textbf{ not ambiguous }\\ Pronoun in place of \texttt{[MASK]}?: \textbf{ no }\\ 
Annotator's answer: \textbf{ Pinheiros }\\ Correct?: \textbf{ yes } 
\item In the next round , Williams faced Alona Bondarenko and once again won with only dropping 5 games . In the quarterfinals , for the third match in a row , \texttt{[MASK]} only dropped five games this time to Czech Lucie  afářová . \\ 
\textbf{ not ambiguous }\\ Pronoun in place of \texttt{[MASK]}?: \textbf{ yes }\\ 
Annotator's answer: \textbf{ Williams }\\ Correct?: \textbf{ yes } 
\item Most traffic is along the stretch between Falkner and Ripley . The Mississippi Department of Transportation calculated an average of 14 , 000 vehicles passing along the route near \texttt{[MASK]} . \\ 
\textbf{ ambiguous }\\ Pronoun in place of \texttt{[MASK]}?: \textbf{ no }\\ 
Annotator's answer: \textbf{ N/A }\\ Correct?: \textbf{ N/A } 
\item Asmodeus kills Vicki and then attacks Dave and Susan . Dave and \texttt{[MASK]} flee to a cemetery and destroy the demon with a cross . \\ 
\textbf{ not ambiguous }\\ Pronoun in place of \texttt{[MASK]}?: \textbf{ no }\\ 
Annotator's answer: \textbf{ Susan }\\ Correct?: \textbf{ yes } 
\item In France , Andrianarivo met with former President Albert Zafy on June 11 , 2007 ; Zafy had also met with Ratsiraka and former Deputy Prime Minister Pierrot Rajaonarivelo in the previous days . \texttt{[MASK]} and Ratsiraka met with Zafy again on June 25 . \\ 
\textbf{ not ambiguous }\\ Pronoun in place of \texttt{[MASK]}?: \textbf{ no }\\ 
Annotator's answer: \textbf{ Andrianarivo }\\ Correct?: \textbf{ yes } 
\item His testimony addressed the key " lie " : that Clinton was allegedly pressuring Betty Currie and Blumenthal himself to attest that it was Lewinsky who initially pursued \texttt{[MASK]} , not vice versa . \\ 
\textbf{ not ambiguous }\\ Pronoun in place of \texttt{[MASK]}?: \textbf{ yes }\\ 
Annotator's answer: \textbf{ Clinton }\\ Correct?: \textbf{ yes } 
\item Leornardo sent Garrett , Vasili Dassiev , Shoji Soma , and Daniel Whitehall to Giza to acquire a power source from a Brood vessel after destroying the Brood inside . \texttt{[MASK]} approved the idea of using the power source to run the rejuvenations chambers found by another team . \\ 
\textbf{ ambiguous }\\ Pronoun in place of \texttt{[MASK]}?: \textbf{ no }\\ 
Annotator's answer: \textbf{ N/A }\\ Correct?: \textbf{ N/A } 
\item Eventually , Saori and Yoshino rejoin Shuichi 's group of friends , though Saori says she still hates Yoshino and Momoko . Shuichi and Anna start dating , much to the surprise of their friends and \texttt{[MASK]} 's sister . \\ 
\textbf{ ambiguous }\\ Pronoun in place of \texttt{[MASK]}?: \textbf{ no }\\ 
Annotator's answer: \textbf{ N/A }\\ Correct?: \textbf{ N/A } 
\item Gmina Branice contains the villages and settlements of Bliszczyce , Boboluszki , Branice , Dzbańce , \texttt{[MASK]} - Osiedle , Dzierżkowice , Gródczany , Jabłonka , Jakubowice , Jędrychowice , Lewice , Michałkowice , Niekazanice , Posucice , Turków , Uciechowice , Włodzienin , Wódka and Wysoka . \\ 
\textbf{ not ambiguous }\\ Pronoun in place of \texttt{[MASK]}?: \textbf{ no }\\ 
Annotator's answer: \textbf{ Dzbańce }\\ Correct?: \textbf{ yes } 
\item When she sees Franky getting into Luke 's car , she gets into a van with Matty . Grace joins him , wanting to talk , but Liv rushes up and demands they follow Franky and \texttt{[MASK]} . \\ 
\textbf{ not ambiguous }\\ Pronoun in place of \texttt{[MASK]}?: \textbf{ no }\\ 
Annotator's answer: \textbf{ Luke }\\ Correct?: \textbf{ yes } 
\item She explained that Laura was frantic and told her that Luis jumped in the water channel and that she was unable to see him anymore . Supposedly , the group of friends met \texttt{[MASK]} at the park and started looking for Luis . \\ 
\textbf{ not ambiguous }\\ Pronoun in place of \texttt{[MASK]}?: \textbf{ yes }\\ 
Annotator's answer: \textbf{ Laura }\\ Correct?: \textbf{ yes } 
\item Whether it was Sidney who next challenged  Vere to a duel or the other way around ,  Vere did not take it further , and the Queen personally took Sidney to task for not recognizing the difference between his status and  \texttt{[MASK]} 's . \\ 
\textbf{ not ambiguous }\\ Pronoun in place of \texttt{[MASK]}?: \textbf{ no }\\ 
Annotator's answer: \textbf{ Vere }\\ Correct?: \textbf{ yes } 
\item Wei attempts to rescue Ku , only to find out that Po deduced Wei 's identity as a cop , since \texttt{[MASK]} was too skilled compared to the rest of his gang . \\ 
\textbf{ not ambiguous }\\ Pronoun in place of \texttt{[MASK]}?: \textbf{ yes }\\ 
Annotator's answer: \textbf{ Po }\\ Correct?: \textbf{ no } 
\item After an unsuccessful evening on the town , Clark takes Sarah to the Indian side of Calcutta , where they attend a party at the home of a wealthy socialite . There , \texttt{[MASK]} seduces Sarah by challenging her to taste life . \\ 
\textbf{ not ambiguous }\\ Pronoun in place of \texttt{[MASK]}?: \textbf{ yes }\\ 
Annotator's answer: \textbf{ Clark }\\ Correct?: \textbf{ yes } 
\item The incapacitated Mike is stabbed repeatedly by Erin , who ties a rope around his neck , attaches the other end to a tractor , and drives the vehicle until Mike 's neck snaps . \texttt{[MASK]} stumbles outside , and discovers Danny , who is barely alive . \\ 
\textbf{ not ambiguous }\\ Pronoun in place of \texttt{[MASK]}?: \textbf{ yes }\\ 
Annotator's answer: \textbf{ Erin }\\ Correct?: \textbf{ yes } 
\item Slingsby was married to the sister of Lawford 's wife , hence why Lawford had to give Slingsby a chance to command the Light Company to prove himself which angered Sharpe . \texttt{[MASK]} was regarded as a poor officer who was often drunk . \\ 
\textbf{ ambiguous }\\ Pronoun in place of \texttt{[MASK]}?: \textbf{ no }\\ 
Annotator's answer: \textbf{ N/A }\\ Correct?: \textbf{ N/A } 
\item Hans and Gerda 's mutual attraction is a challenge , as Gerda is navigating her changing relationship to Lili ; but Hans ' long - time friendship with and affection for \texttt{[MASK]} cause him to be supportive of both Lili and Gerda . \\ 
\textbf{ ambiguous }\\ Pronoun in place of \texttt{[MASK]}?: \textbf{ no }\\ 
Annotator's answer: \textbf{ N/A }\\ Correct?: \textbf{ N/A } 
\item Sirius " sailed in ballast , having unloaded a cargo of hay at Røsneshavn after departing Tromsø . She had left \texttt{[MASK]} in the morning of 17 May 1940 . " \\ 
\textbf{ not ambiguous }\\ Pronoun in place of \texttt{[MASK]}?: \textbf{ no }\\ 
Annotator's answer: \textbf{ Røsneshavn }\\ Correct?: \textbf{ yes } 
\item Harold does not take the news well , but Karl eventually convinces him to fight . \texttt{[MASK]} has an operation and begins chemotherapy after speaking to Stephanie Scully . \\ 
\textbf{ not ambiguous }\\ Pronoun in place of \texttt{[MASK]}?: \textbf{ yes }\\ 
Annotator's answer: \textbf{ Harold }\\ Correct?: \textbf{ yes } 
\item Ron knocks Dale out and leaves her in a locked car filling with exhaust , sadistically goading Andrew into braving his agoraphobia in order to save her . Andrew manages to save her and wound Ron ; reviving , Dale deals \texttt{[MASK]} a death blow . \\ 
\textbf{ not ambiguous }\\ Pronoun in place of \texttt{[MASK]}?: \textbf{ no }\\ 
Annotator's answer: \textbf{ Ron }\\ Correct?: \textbf{ yes } 
\item Evans introduced two romantic interests for Corrigan : Anina Kreemar , the wealthy niece of Corrigan 's bureau chief , and \texttt{[MASK]} 's friendly rival Jennever Brand , a spirited female agent of a rival clandestine spy agency . \\ 
\textbf{ not ambiguous }\\ Pronoun in place of \texttt{[MASK]}?: \textbf{ yes }\\ 
Annotator's answer: \textbf{ Corrigan }\\ Correct?: \textbf{ yes } 
\item Her fifth victim was Pillama , aged 60 , and killed at Maddur Vyadyanathapura . \texttt{[MASK]} was a temple priest at Hebbal temple . \\ 
\textbf{ not ambiguous }\\ Pronoun in place of \texttt{[MASK]}?: \textbf{ yes }\\ 
Annotator's answer: \textbf{ Pillama }\\ Correct?: \textbf{ yes } 
\item Sandy tells Rizzo she plans to watch the race and offers to help \texttt{[MASK]} despite the rumors about Rizzo 's character that have been spread around school . \\ 
\textbf{ not ambiguous }\\ Pronoun in place of \texttt{[MASK]}?: \textbf{ yes }\\ 
Annotator's answer: \textbf{ Rizzo }\\ Correct?: \textbf{ yes } 
\item Qianru 's mother , Fengyi tries to talk her into accepting the fact that Huanhuan is autistic but Qianru is unwilling to face reality . After some time , under Wenxin 's patient persuasion , \texttt{[MASK]} finally agrees to send Huanhuan to a school for children with special needs . \\ 
\textbf{ not ambiguous }\\ Pronoun in place of \texttt{[MASK]}?: \textbf{ yes }\\ 
Annotator's answer: \textbf{ Qianru }\\ Correct?: \textbf{ yes } 
\item After Trey pushes away Guy , Trey finally realized that Alex was right all along and that Guy has been trying to break them up . Trey and Alex kick \texttt{[MASK]} out of their home and later apologizes to Alex . \\ 
\textbf{ not ambiguous }\\ Pronoun in place of \texttt{[MASK]}?: \textbf{ yes }\\ 
Annotator's answer: \textbf{ Guy }\\ Correct?: \textbf{ yes } 
\item Burns 's clashes with Smith was perhaps most obvious at the notorious New York City concert in 1998 where Burns attacked \texttt{[MASK]} after the vocalist repeatedly and deliberately knocked one of Burns 's cymbal stands to the floor . \\ 
\textbf{ not ambiguous }\\ Pronoun in place of \texttt{[MASK]}?: \textbf{ yes }\\ 
Annotator's answer: \textbf{ Smith }\\ Correct?: \textbf{ yes } 
\item When Viki asks if he loves Echo , Charlie hesitates , and Viki storms off . On April 12 , Viki asks \texttt{[MASK]} again whether or not he loves Echo ; he says he does . \\ 
\textbf{ not ambiguous }\\ Pronoun in place of \texttt{[MASK]}?: \textbf{ yes }\\ 
Annotator's answer: \textbf{ Charlie }\\ Correct?: \textbf{ yes } 
\item In pre - sentence proceedings , Chen 's father , Edward Chen , was reported as saying : During his final plea on 2 February 2006 , Chen said : On 15 February 2006 \texttt{[MASK]} was sentenced to life imprisonment . \\ 
\textbf{ not ambiguous }\\ Pronoun in place of \texttt{[MASK]}?: \textbf{ yes }\\ 
Annotator's answer: \textbf{ Chen }\\ Correct?: \textbf{ yes } 
\item Andrea considers assassinating the Governor , but Milton knows that his second - in - command , Martinez , will follow through on the Governor 's plans . Instead , \texttt{[MASK]} urges Andrea to escape and warn Rick and the others . \\ 
\textbf{ not ambiguous }\\ Pronoun in place of \texttt{[MASK]}?: \textbf{ yes }\\ 
Annotator's answer: \textbf{ Milton }\\ Correct?: \textbf{ yes } 
\item Patterson took the communiqué to the White House , where Truman and Attlee signed it on 16 November 1945 . The next meeting of the Combined Policy Committee on 15 April 1946 produced no accord on collaboration , and resulted in an exchange of cables between Truman and \texttt{[MASK]} . \\ 
\textbf{ ambiguous }\\ Pronoun in place of \texttt{[MASK]}?: \textbf{ no }\\ 
Annotator's answer: \textbf{ N/A }\\ Correct?: \textbf{ N/A } 
\item In any case , Baldwin 's other brother Philip of Namur remained as regent , and eventually both of \texttt{[MASK]} 's daughters , Joan and Margaret II , were to rule as countesses of Flanders . \\ 
\textbf{ not ambiguous }\\ Pronoun in place of \texttt{[MASK]}?: \textbf{ no }\\ 
Annotator's answer: \textbf{ Baldwin }\\ Correct?: \textbf{ yes } 
\item In the late 1990s , Luis Rossi , Ivan Fernandez , and Mercedes Fernandez purchased the Aragon . In September 2014 , \texttt{[MASK]} sold all her interests in the Aragon . \\ 
\textbf{ not ambiguous }\\ Pronoun in place of \texttt{[MASK]}?: \textbf{ no }\\ 
Annotator's answer: \textbf{ Mercedes Fernandez }\\ Correct?: \textbf{ yes } 
\item Peschko also played chamber music ; best known are his projects with violinist Georg Kulenkampff and cellists Enrico Mainardi and Hans Adomeit . From 1953 to 1958 \texttt{[MASK]} was responsible for lieder , choir and church music at Radio Bremen . \\ 
\textbf{ not ambiguous }\\ Pronoun in place of \texttt{[MASK]}?: \textbf{ yes }\\ 
Annotator's answer: \textbf{ Peschko }\\ Correct?: \textbf{ yes } 
\item Realising the war was lost , Himmler attempted to open peace talks with the western Allies without Hitler 's knowledge , shortly before the end of the war . Hearing of this , \texttt{[MASK]} dismissed him from all his posts in April 1945 and ordered his arrest . \\ 
\textbf{ not ambiguous }\\ Pronoun in place of \texttt{[MASK]}?: \textbf{ yes }\\ 
Annotator's answer: \textbf{ Hitler }\\ Correct?: \textbf{ yes } 
\item Dein was behind the appointment of the then little known Arsène Wenger to the manager 's job in 1996 ; under Wenger , Arsenal have won the Premier League three times and the FA Cup seven times , and \texttt{[MASK]} strongly backed him and his transfer wishes throughout . \\ 
\textbf{ not ambiguous }\\ Pronoun in place of \texttt{[MASK]}?: \textbf{ no }\\ 
Annotator's answer: \textbf{ Dein }\\ Correct?: \textbf{ yes } 
\end{enumerate}

\end{document}